\documentclass[fleqn,10pt,lineno]{wlpeerj} 

\usepackage{apacite}
\usepackage{amsmath}
\usepackage{graphicx, caption, subcaption}
\usepackage{natbib}
\usepackage{url}
\nolinenumbers

\title{Learning to Perform Role-Filler Binding with Schematic Knowledge}

\author[1]{Catherine Chen}
\author[2]{Qihong Lu}
\author[2]{Andre Beukers}
\author[3]{Christopher Baldassano}
\author[2]{Kenneth A. Norman}
\affil[1]{Department of Computer Science, Princeton University, Princeton, New Jersey, USA}
\affil[2]{Department of Psychology, Princeton University, Princeton, New Jersey, USA}
\affil[3]{Department of Psychology, Columbia University, New York, New York, USA}
\corrauthor[1]{Catherine Chen}{cc27@alumni.princeton.edu}
\newcommand*{\storyfont}{\fontfamily{pcr}\selectfont}
\DeclareTextFontCommand{\textstoryfont}{\storyfont}
\newcommand{\figurepath}{./figures/}

\begin{abstract}
\nolinenumbers

    Through specific experiences, humans learn relationships underlying the structure of events in the world. Schema theory suggests that we organize this information in mental frameworks called ``schemata," which represent our knowledge of the structure of the world. Generalizing knowledge of structural relationships to new situations requires role-filler binding, the ability to associate specific ``fillers" with abstract ``roles." For instance, when we hear the sentence ``Alice ordered a tea from Bob," the role-filler bindings ``Alice:customer," ``tea:drink," and ``Bob:barista" allow us to understand and make inferences about the sentence. We can perform these bindings for arbitrary fillers -- we understand this sentence even if we have never heard the names ``Alice," ``tea," or ``Bob" before.
    In this work, we define a model as capable of performing role-filler binding if it can recall arbitrary fillers corresponding to a specified role, even when these pairings violate correlations seen during training. Previous work found that models can learn this ability when explicitly told what the roles and fillers are, or when given fillers seen during training. We show that networks with external memory can learn these relationships with fillers not seen during training and without explicitly labeled role-filler bindings, and show that analyses inspired by neural decoding can provide a means of understanding what the networks have learned.
\end{abstract}

\begin{document}
\bibliographystyle{plainnat}
\setcitestyle{notesep={; },round,aysep={},yysep={;}}
\flushbottom
\maketitle
\thispagestyle{empty}

\section*{Introduction}
Knowing how events are structured in the world allows humans to understand and interact with novel situations. As humans, we have a powerful ability to learn the relationships underlying the structure of events, and to use them to organize and guide cognition. Schema theory suggests that mental frameworks called ``schemata" organize our knowledge of the world. Humans learn schemata through experiences and use them as building blocks for understanding the world. For example, we learn the schema for ``visiting coffee shops" based on individual experiences at specific coffee shops. Although each coffee shop visit differs slightly from the others, the experiences share some underlying structure, which we are able to learn without being explicitly instructed about the underlying structural relationships. Previous work studying these phenomena has referred to these structures as ``scripts" \citep[e.g.][]{SchankAbelson:1977,Bower:1979,Miikkulainen:1991} and ``frames" \citep[e.g.][]{Minsky:1974,Brachman:1985}, and has shown that they affect people's memory of events \citep{Bartlett:1932,Bower:1979}.

These structures can be viewed as frames consisting of abstract ``roles" which are occupied by specific ``fillers" \citep{Minsky:1974,Brachman:1985}. For instance, our schema for coffee shops might include the roles ``barista," ``drink," and ``customer." Knowing the coffee shop schema allows us to understand and make predictions based on the sentence \textit{Alice ordered a green tea from Bob}, by inferring the role-filler relations \textit{barista:Bob}, \textit{drink:green tea}, \textit{customer:Alice}. We can do this even if we have no idea what the words ``Alice," ``green tea," and ``Bob" mean. This kind of inferential process critically relies on an operation that binds a specific filler (e.g. ``Alice") to a known structural ``role" (e.g. ``customer"). This process is commonly referred to as ``role-filler binding." Role-filler binding is essential for understanding and organizing structural relationships within the world, allowing us to learn flexible, composable building blocks with which to understand new situations.

Role-filler binding involves the ability to systematically apply propositions to new fillers. It allows us to generalize schematic knowledge to novel instances, by applying the relationships in a schema to arbitrary fillers. For instance, if we understand the phrase ``Alice ordered a green tea from Bob," we should be able to understand the phrase ``A ordered a C from B" for any fillers ``A," ``B," and ``C." Previous work has stated that systematicity requires bindings that are independent of the fillers they bind, and which can dynamically bind different fillers to different roles in each situation \citep{HolyoakHummel:2000}. Therefore we test whether networks can recall arbitrary fillers from examples that fit the structure they are trained on, even when these pairings violate correlations seen during training. If they can, then we interpret the model as capable of performing role-filler binding.

We test whether connectionist architectures can perform role-filler binding. We provide models with examples generated from a certain statistical structure, and train them to recall the fillers corresponding to certain roles.

We find a model and training regime that result in successful role-filler binding. This occurs with words that were never seen during training, and with inputs that do not explicitly encode role-filler pairs. We characterize architecture- and task- boundaries of this ability by finding models that are \textit{not} sufficient for role-filler binding, and inputs with structures that heavily diverge from structures seen during training, that our model and training combinations do not solve.

Specifically, we show that some connectionist architectures (the Fast Weights and reduced Neural Turing Machine) are able to perform role-filler binding without explicitly labeled roles and fillers, but not all neural networks do this. We find that models must see a sufficiently diverse set of training examples in order to perform schema-generalization, and that this model/training combination fails when the schema changes too much between train and test examples. Lastly, we provide additional analyses inspired by neural decoding that give insights into the how successful models solve this task.

\section*{Related Work}
Previous work has pointed to the importance of being able to represent role-filler bindings independently and dynamically, such that roles can be bound to different fillers in different scenarios \citep[e.g.][]{HummelBiederman:1992,DoumasHummel:2005}.
Connectionist models were traditionally thought to be unable to represent these symbolic relations, because of an ability to independently represent roles and fillers \citep[e.g.][]{FodorPylyshyn:1988,DoumasHummel:2005}.

Prior work has shown that neural networks can learn role-filler binding if they are given inputs that explicitly provide the role corresponding to each filler, or if they are given train and test inputs that share the same pool of fillers. For instance, some models use specific input units to encode roles such as the action and agent of the example \citep{Kriete:2013,Elman:2019}, and others use holographic reduced representations \citep{Plate:1995} to explicitly encode the action, agent, and patient of the input examples \citep{Franklin:2019}. Another line of work found models that could perform role-filler binding when tested on examples containing the same pool of fillers used during training \citep{StJohnMcClelland:1990,Miikkulainen:1991,Hinaut:2013}. In some cases, these models can generalize to unseen grammatical constructions that are compositions of trained constructions, if the examples contain the same set of fillers during train and test \citep{Hinaut:2013}. Others have developed models that represent inputs as combinations of features, and presented a hand-crafted sequence of comparison and binding operations that allow the model to perform dynamic role-filler binding \citep{Doumas:2008}.

Recent work tested role-filler binding, testing models' ability to perform role-filler binding on previously unseen fillers, role-filler bindings that violate statistical correlations seen during training, and examples where the input segments are presented in a shuffled order \citep{Puebla:2019}. This work suggested that connectionist architectures may not be able to learn relational structures necessary to perform role-filler binding, showing that a Story Gestalt \citep{StJohnMcClelland:1990} and a Seq2Seq with Attention \citep{Bahdanau:2015} model fail to perform role-filler binding when given novel fillers or bindings that violate correlations seen during training. Our work builds upon these findings, by identifying models that learn to perform role-filler binding in a way that generalizes to novel fillers, and to role-filler bindings that violate correlations seen during training.

Our decoding analyses are inspired by multivariate pattern analyses used to decode neural data \citep{Norman:2006}. Rather than neuroimaging data, we apply these methods to neural network activity. Previous researchers have used various mapping methods to gain insight into neural network activity based on the activation of network layers. For instance, previous work has used stimulus-decoding analyses and activation similarity to probe for features represented by networks and to gain insight into the processing stages corresponding to layers of the networks \citep[e.g.][]{Ettinger:2016,Qian:2016,Hupkes:2018,GuestLove:2019,Lakretz:2019,Tenney:2019}.

\section*{Methods}
We use schematically generated stories to test networks' ability to perform role-filler binding. We generated stories using the Coffee Shop World, which takes as input an underlying graph that defines story states and transition probabilities between states, and produces stochastically generates stories \citep{csw}. Each state includes fixed frame-text and variable roles, and the roles are substituted with fillers drawn from a specified pool in each instance of the story. For example, consider the state \textit{Order\_food}:

\begin{center}
\textit{[subject] ordered a plate of [dessert]}
\end{center}
In a specific instance of the state, the roles \textit{subject} and \textit{dessert} would be occupied by randomly chosen fillers, such as ``Alice" and ``chocolate."

\begin{figure}[htb!]
\centering
\includegraphics[width=0.7\linewidth]{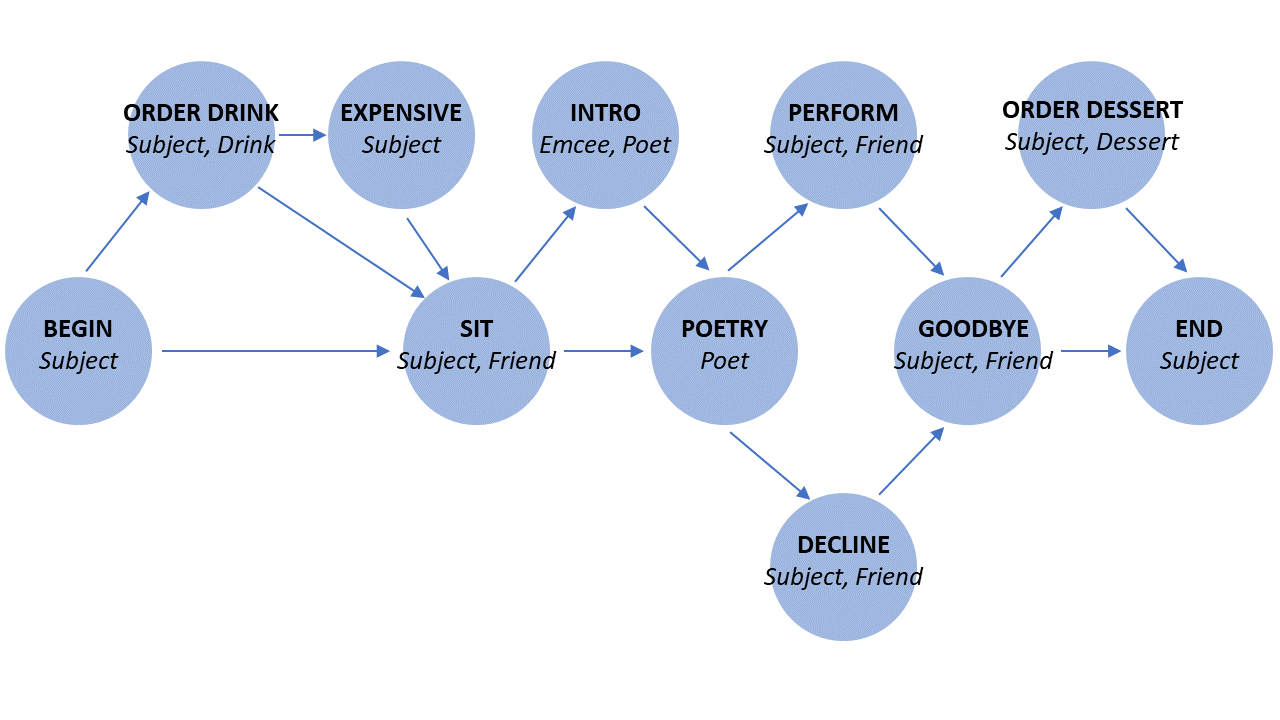}
\caption{\textbf{Story Graph for Role-Filler Binding Experiments.} Each edge indicates a possible transition. For states with multiple outgoing transitions, each outgoing transition is equally likely.}\label{fig:story_paths}
\end{figure}

In Figure \ref{fig:story_paths} and Table \ref{tab:story_paths}, we show the specific schema used in our experiments.

\begin{table}[htb!]
\centering
    \includegraphics[width=0.6\linewidth]{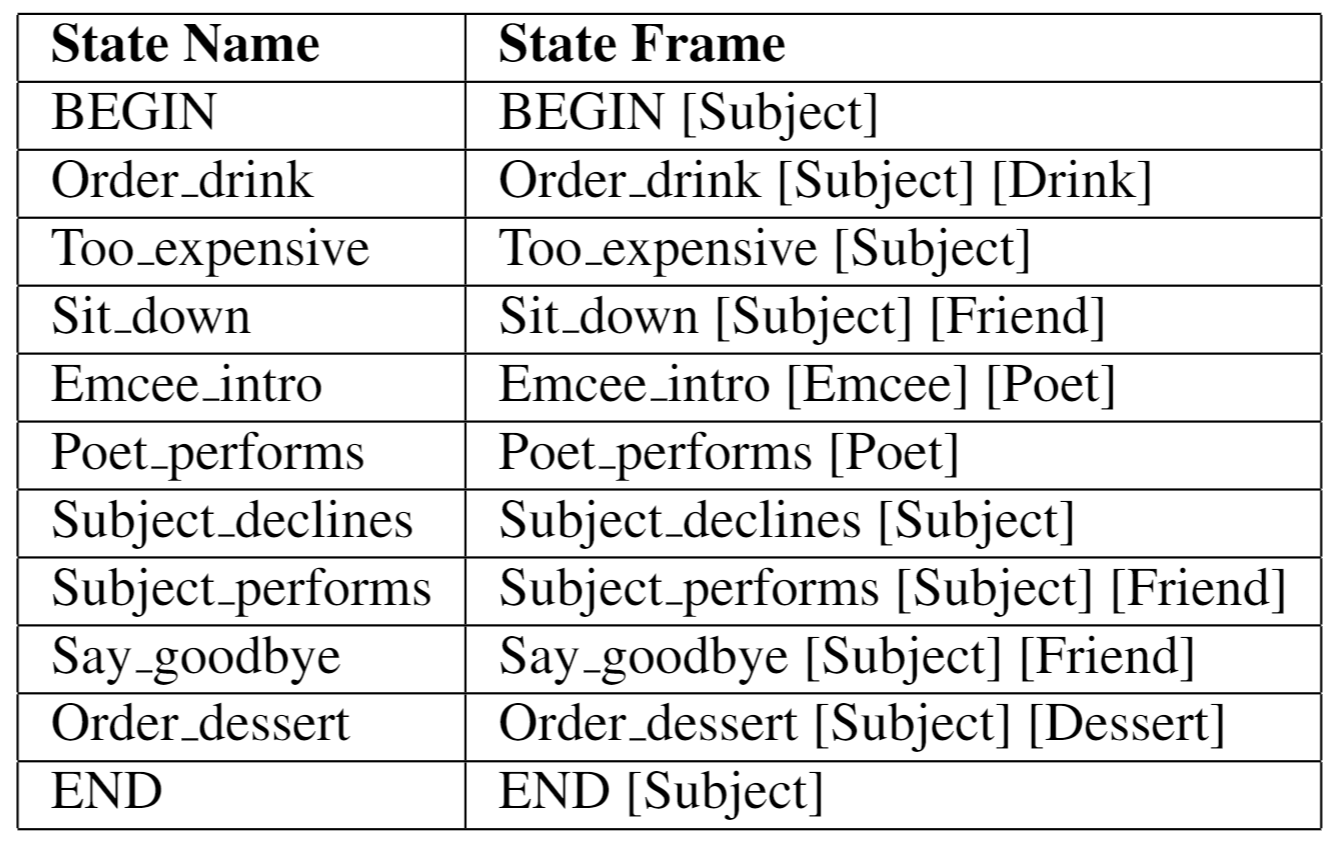}
	\caption{\textbf{Story States for Role-Filler Binding Experiments.} We provide the text of each state of the story, where the bracketed roles are substituted by specific fillers in each story.}\label{tab:story_paths}
\end{table}

\subsection*{Tasks}
In our tasks we presented networks with stories generated by Coffee Shop World, and then queried the networks for the filler corresponding to a specified role. For instance, the network might receive the input
\begin{quote}
\textit{begin alice sit alice bob poet\_performs chris subject\_performs alice bob say\_goodbye alice bob end alice qpoet}
\end{quote}

In this case the correct output is ``chris," because ``chris" is the filler corresponding to the role ``poet" in the schema we define in Figure \ref{fig:story_paths} and Table \ref{tab:story_paths}.

We train the network on a certain pool of fillers, and construct test sets in which inputs contain fillers not seen during training. Furthermore, the position of the filler corresponding to a given role is not necessarily the same in each story because transitions between states are probabilistic. Therefore successfully complete these tasks, the networks cannot simply memorize a word or word position to answer each query. The networks must learn to extract the filler corresponding to each role, store these role-filler pairs during the input sequence, and select the correct filler to output after receiving the query.

We show that there exist networks that can learn role-filler binding (rather than proving that some specific network architecture performs best on a downstream language processing task). Therefore the importance of our evaluation metrics is to show that networks are capable of learning the task, and we do not compare hyperparameters or sample efficiency in this work.

\subsection*{Input Representations}
We represented each word of the input as a randomly generated 50-dimensional vector. Each index of the vector is independently drawn from a $N(0,1)$ distribution, and then the vectors are normalized to have unit Euclidean norm. We use randomly generated vectors so that networks could not extract hints about schematic structure from information encoded in pre-trained word vectors. We sequentially fed the words of the story into the networks, followed by a query word indicating which filler to retrieve. The network then outputted a 50-dimensional vector, and we computed the cosine similarity between this prediction and each vector in the experiment's corpus, choosing the most similar word as the network's prediction.

We inserted a nonsense word (a randomly generated vector that does not represent any other word in the corpus) into a randomly chosen location in the input story, to force the network to learn representations of the schema that are robust to small position shifts.

To ensure that all inputs have the same number of words, we padded inputs with nonsense words between the end of a story and the appearance of the query.

\subsection*{Training Regimes}
We ran experiments with two types of training regimes: Limited Filler Training and Unlimited Filler Training. In ``Limited Filler Training" experiments, we substituted roles with fillers drawn from a small, finite pool of fillers (with six possible fillers for each role). Within this category of experiments, we tested on previously seen and previously unseen fillers. When testing with previously seen fillers, the pool of fillers was the same during training and testing. When testing with previously unseen fillers, we drew from disjoint pools of fillers during training and testing, meaning that the network needed to perform role-filler binding with fillers it had never seen before. In all experiments, we ensured that the train and test set contained distinct input sequences (i.e. they could not contain inputs with both the same sequence of states and the same role-filler pairs).

In ``Unlimited Filler Training" experiments, we randomly generated a new vector for each filler for each input story during both training and testing, rather than using a finite pool of fillers. In this case, during both training and testing, the network was continuously asked to perform role-filler binding using previously unseen fillers.

\subsubsection*{Prediction Method}
To determine the network's prediction, we used networks in which the final layer has $50$ nodes. We computed the cosine similarity between the output vector and the vector embedding of each word in the experiment's corpus, and selected the word with the highest cosine similarity to the network's output vector.

The set of possible words is the corpus created by combining the words seen in all stories in a particular training batch. For fixed embeddings, the corpus therefore consists of all of the words that occur in the stories generated for a particular experiment. For experiments in which we generated a new random embedding for each story, the corpus also includes all the filler vectors newly generated for stories in that particular batch.

\subsubsection*{Chance Rates}
In each experiment, the network's chance rate depends on the number of words it has to choose from.

In Experiments $1$ (Limited Filler Training, tested with previously seen fillers) and $2$ (Limited Filler Training, tested with previously unseen fillers), the network must choose from a corpus size of $44$, corresponding to a chance rate of $2.3\%$.

In Experiment $3$ (Unlimited Filler Training, tested with previously unseen fillers) the network must choose from all the words in the story corpus ($25+n$, where $n$ is the number of possible queries) and the newly generated representations for each story in the batch. Since we use a validation batch size of $4$, and $12$ new filler vectors are generated for each input, this results in a total of between $25+1+4\times 12=75$ and $31+1+4\times 12=80$ words, for a chance rate of around $1.3\%$.

\subsubsection*{Epoch Sizes}
In Experiment $1$ we used $47135$ train and $11784$ test stories. In Experiment $2$ we used $55448$ train and $3339$ test stories. These numbers were the result of generating $100000$ stories using Coffee Shop World \cite{csw} and removing any repeated stories. In Experiment 1, $80\%$ of stories were used for training and the remaining $20\%$ used for testing. In Experiment 2, the stories were partitioned to ensure that there was no overlap between fillers used during testing and training. 

In Experiment $3$ we used $112$ train and $112$ test stories. To compute the number of distinguishable stories for this experiment we summed over the number of possible queries (queries that can be answered using the information in the story; for instance, some stories may not include an \textit{Emcee} and therefore the input must not use \textit{QEmcee} as a task) for each possible traversal through the story graph. This gives us $112$ stories.

\subsection*{Models}
We tested four recurrent neural network (RNN) architectures. RNNs are a class of neural network architectures with weights that form directed cycles. The cycles form feedback loops that allow networks to maintain an internal state. The structure of RNNs allowed us to provide the input story one word at a time, followed by the query. The feedback loops allowed for a form of short-term memory (where we define ``short-term" as the timescale of a single story), with which they could maintain relevant parts of the story. We tested multiple RNN network architectures, to investigate which memory components (if any) are sufficient for role-filler binding. For each of our architectures we used $50$ hidden units and a learning rate of $1e-4$.

In addition to a standard RNN, we tested Long Short-Term Memory (LSTM), Fast Weights, and reduced Neural Turing Machine (NTM) architectures. We used layer normalization for the RNN, LSTM, and Fast Weights architectures, which re-centers and re-scales the networks' layers and serves to stabilize the network dynamics \citep{Ba:2016a}.

The LSTM consists of an RNN with gates to control what the internal state stores, forgets, and displays to the rest of the network \citep{Hochreiter:1997}. The Fast Weights architecture consists of an RNN with a matrix of quickly changing ``fast weights" \citep{Ba:2016b}. This extra matrix of weights allows for auto-associative memory, and the combination of the quickly changing fast weights matrix and more slowly changing standard weights, is inspired by different speeds of change in biological neuronal connections \citep{Martin:2000}. The NTM is an RNN with an LSTM ``controller" that learns to read to and write from an external buffer \citep{Graves:2014}. The network can use an external buffer as a ``mental scratchpad" to store and retrieve memories, and the controller must learn how to use this external buffer. The combination of a controller and external buffer is inspired by interactions between the hippocampus and cortex in the human brain, which play a key role in human memory \citep{Oreilly:2014}. We use a reduced NTM, which we construct by removing the links between adjacent external buffer slots in a standard NTM. These links were originally included to allow the network to learn associations between external memories. Role-filler binding should not require the network to maintain links between these memories, since each role-filler binding should be independent of the others. Removing these links reduces the number of parameters our model needs to learn, and clarifies the analyses in our decoding experiments. Our reduced NTM model has a memory size of $128$, a word size of $20$, $1$ write head, and $4$ read heads.

These networks have shown success on a range of tasks including speech recognition \citep{Graves:2013}, language modeling \citep{Mikolov:2012}, and associative recall \citep{Graves:2014}, but to our knowledge, none have been shown to learn a representation of a schematic structure that extends to previously unseen fillers, without explicit labeling of roles and fillers.

\subsection*{Decoding Analysis}
We performed decoding analyses to analyze the mechanisms underlying task performance.

For each model, we recorded network activity after the model received each word in an input sequence. For each example and each model, this resulted in one 50-dimensional vector of hidden unit activity per input word. The Fast Weights and NTM networks consist of two memory components (the hidden state and an external memory component), and we additionally recorded the values of the external memory components. From the Fast Weights network, we obtained a 50-dimension vector of hidden unit activity, and a 50-by-50-dimension matrix (which we flatten into a 2500-dimension vector) of fast weights activity, after each word in the input sentence. From the reduced NTM, we obtained a 50-dimensional vector of hidden unit activity, and a 128-by-20-dimension matrix (corresponding to 128 memory slots each of size 20, which we flatten into a 2560-dimension vector) of memory buffer activity, after each word in the input sentence.

We constructed 100 input sequences with the same story frame and completed each sequence with distinct fillers (i.e. the frame text for each sequence was identical, but the fillers were different). For each role, we trained a ridge regression mapping (with regularization stringth $1.0$) between each memory state vector and correct output fillers, using 80 of the sequences for training. For each role, this resulted in six regression mappings: four mappings from each model's 50-dimension hidden state to the 50-dimension correct output filler, one mapping from the 2500-dimension vector of fast weights to the 50-dimension correct output fillers, and one mapping from the 2560-dimension reduced NTM memory buffer to the 50-dimension correct output fillers. Then on each of the remaining 20 sequences, we used this mapping to predict the output filler. We ranked each corpus vector in terms of its cosine similarity with the predicted output, and computed the ranking score ($1-\frac{\text{rank of actual output}}{\text{corpus size}}$) for each test sequence. These ranking scores have a maximum score of $1$, with a chance rate of $0.5$.

\subsection*{Correlation Violation Tests}
Previous work suggests that connectionist networks learn statistical correlations rather than relational structure; for example, a recent study showed that network performance fell below chance when the network was given test examples with different statistical structure from training examples \citep{Puebla:2019}.
To assess whether this was true of the networks used here, we constructed two additional tasks to test the flexibility of networks' schema-generalization.

The first task probed whether networks can perform role-filler binding on test examples that break correlations observed during training. Networks were trained on stories constructed from the frame ``\textit{begin subject sit subject friend announce emcee perform poet consume dessert drink goodbye.}" In this experiment, the fillers for ``subject," ``friend," ``emcee," and ``poet" were drawn from a pool of fillers, while all other words stayed constant in each example. There were $1000$ possible fillers, and during training each role could be filled by a fixed subset of $750$ of these fillers (so each filler was excluded from one of these four roles during training). During test, each role was filled only by fillers that were excluded from that role during training, to test whether networks learned a strategy for role-filler binding that was robust to violations of correlations seen during training.

Humans are able to generalize role-filler binding to arbitrary fillers that violate correlations seen during training, and also retain knowledge of the correlations seen during training. For instance, if given the sentence `lkw gave a coffee to Bob', we are able to associate the filler `lkw' with the role `barista', even though `lkw' violates the statistical correlations we have observed in our experiences with people's names. Although we can override these statistical correlations when necessary, we retain knowledge of these correlations (e.g. that `Alice' and `Bob' are statistically more likely to be names). We therefore tested whether networks can both accurately extract role-filler bindings that violate correlations seen during training, and retain knowledge of the statistical correlations seen during training. 

To test networks' retention of statistical correlations seen during training, we construct ambiguous input sentences that contain insufficient information for determining the correct response, following \cite{StJohnMcClelland:1990}.
We trained networks on the correlation-inducing training set described above, with two variants. In the first variant, each index of each word vector is independently drawn from a $N(0,1)$ distribution, and then the vector is normalized to have unit Euclidean norm. In the second variant, each index of each word vector is initially drawn from a $N(0,1)$ distribution and the vector is normalized to have unit Euclidean norm, and then for $50\%$ of word vectors representing filler words we add $0.5$ to the even indices of the word vectors. We then constructed an additional set of test inputs, which consist solely of nonsense words (words that the network is trained to ignore, as described in the ``Input Representations" section), followed by a query seen during training (one of \textit{QEmcee}, \textit{QFriend}, \textit{QPoet}, \textit{QSubject}). These ambiguous test examples, used in addition to test examples consisting of correlation-violating inputs, allow us to test both the network's ability to generalize to examples that violate correlations seen during training, and the network's ability to retain and draw upon correlations seen during training.

The second task probes whether networks can generalize to changes in the story structure. Networks were trained on the story ``\textit{begin subject sit subject friend announce emcee perform poet consume dessert drink goodbye}," and tested on a shuffled version of the story: \textit{``consume dessert drink goodbye begin subject sit subject friend announce emcee perform poet."}

\subsection*{Code.}
We used TensorFlow to implement our experiments, adapting existing architecture implementations from (Mohandas, 2018) and (deepmind, 2018).

We used Coffee Shop World to generate the stories used in this experiment. This generator is available on GitHub (\textit{Coffee Shop World}, 2017).

The code used to generate data, run experiments, and generate the plots in this paper is available on GitHub at \url{https://github.com/cchen23/generalized_schema_learning/}. We also include pre-generated data and checkpoints of trained networks.

\section*{Results}
\subsection*{Experiment 1: Limited Filler Training, Tested with Previously Seen Fillers}
In the Limited Filler Training experiment, test inputs used the same pool of fillers as train inputs. During test, the networks were provided with new stories -- they had seen each word of each input before, but never with this particular permutation of words. This experiment tested whether networks possess the ability to learn associations between roles and fillers, given stories generated from an underlying schema.

\begin{figure}[htb!]
\centering
\includegraphics[width=0.7\linewidth]{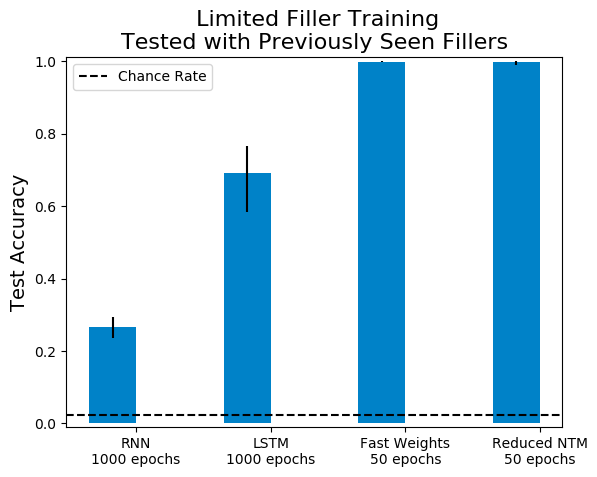}
\caption{\textbf{Test Scores for Experiment 1.} Each architecture is able to learn to perform role-filler binding on a story it has not previously seen, when it has encountered each of the story's words during training. The chance accuracy rate is 2.3\%, bars denote mean training accuracies, and error bars denote maximum and minimum accuracies over three trials. Full learning curves are available in the Supplemental Material.}\label{fig:limitedtrain_previouslyseen_barchart}
\end{figure}

As we show in Figure \ref{fig:limitedtrain_previouslyseen_barchart}, each architecture performs the experiment task at a significantly above chance rate. While the basic RNN learns more slowly than other networks, its architecture is sufficient to learn and apply a schema to situations in which it has seen the fillers before, in a slightly different context.

\subsection*{Experiment 2: Limited Filler Training, Tested with Previously Unseen Fillers}
We conducted a second Limited Filler Training experiment, in which the networks were tested on stories containing fillers they had not encountered during training. This experiment tested whether networks could not only learn to perform role-filler binding, but also generalize their schematic knowledge to fillers they had never encountered.

All networks fail to do so. While all networks perform far above chance on previously seen stories presented during training, the test accuracy of each network remained at $0\%$ (the chance accuracy rate from guessing random vectors is 2.3\%). The test accuracy lies below the chance accuracy rate because the networks overfit to the specific fillers seen during training. When we examined the specific words predicted by the networks, we found that the networks always predict fillers from the training set. Since, in this experiment, the train and test fillers are drawn from disjoint pools, the networks always predict the wrong response during test.

\subsection*{Experiment 3: Unlimited Filler Training, Tested with Previously Unseen Fillers}
We conducted an Unlimited Filler Training experiment, testing the networks with previously unseen fillers. We constructed train and test sets in which fillers were represented by new randomly generated fillers in each example; thus, the networks needed to generalize to previously unseen fillers to succeed in both the train and test sets.

\begin{figure}[!htb]
\centering
\begin{subfigure}[t]{0.4\linewidth}
{\includegraphics[width = \linewidth]{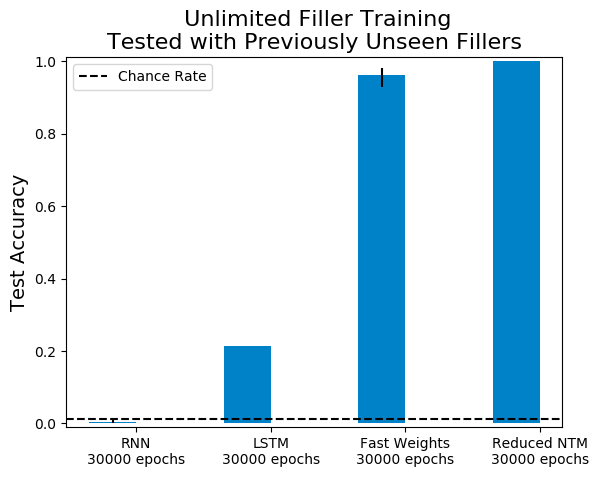}}
\caption{Overall Accuracies. Three architectures reach above-chance test accuracy, showing that certain networks perform some amount of generalization when forced to do so during training.}
\label{fig:Unlimitedtrain_barchart_overall}
\end{subfigure}\hspace{5mm}
\begin{subfigure}[t]{0.4\linewidth}
{\includegraphics[width = \linewidth]
{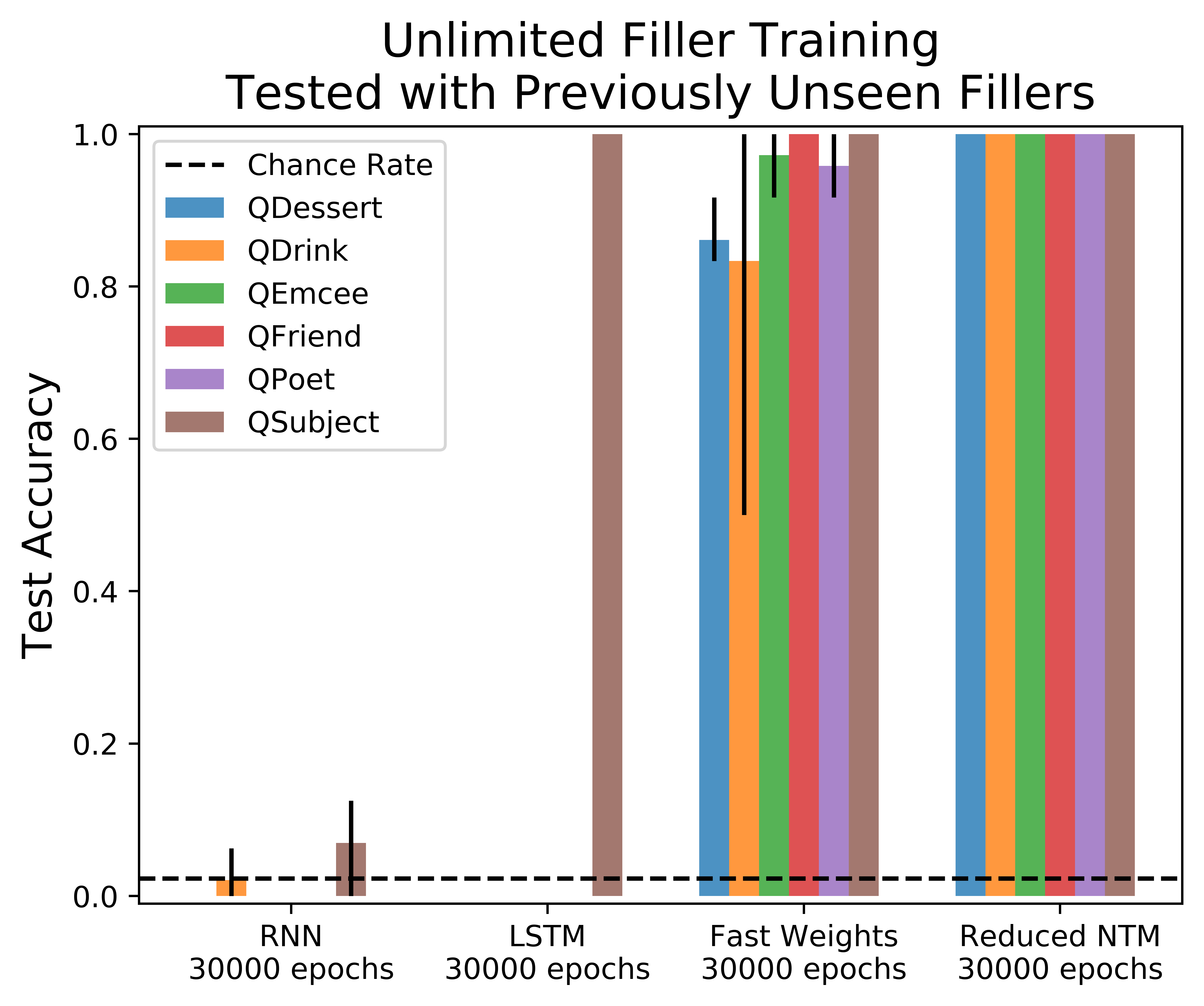}}
\caption{Query-Split Accuracies. The LSTM and RNN learn to generalize only on the \textit{QSubject} task. The reduced NTM and Fast Weights networks learn to solve all six tasks; moreover, they learn to solve simpler tasks more quickly.}
\label{fig:Unlimitedtrain_barchart_split}
\end{subfigure}
\begin{subfigure}[b]{0.1\linewidth}
{\includegraphics[width = \linewidth]{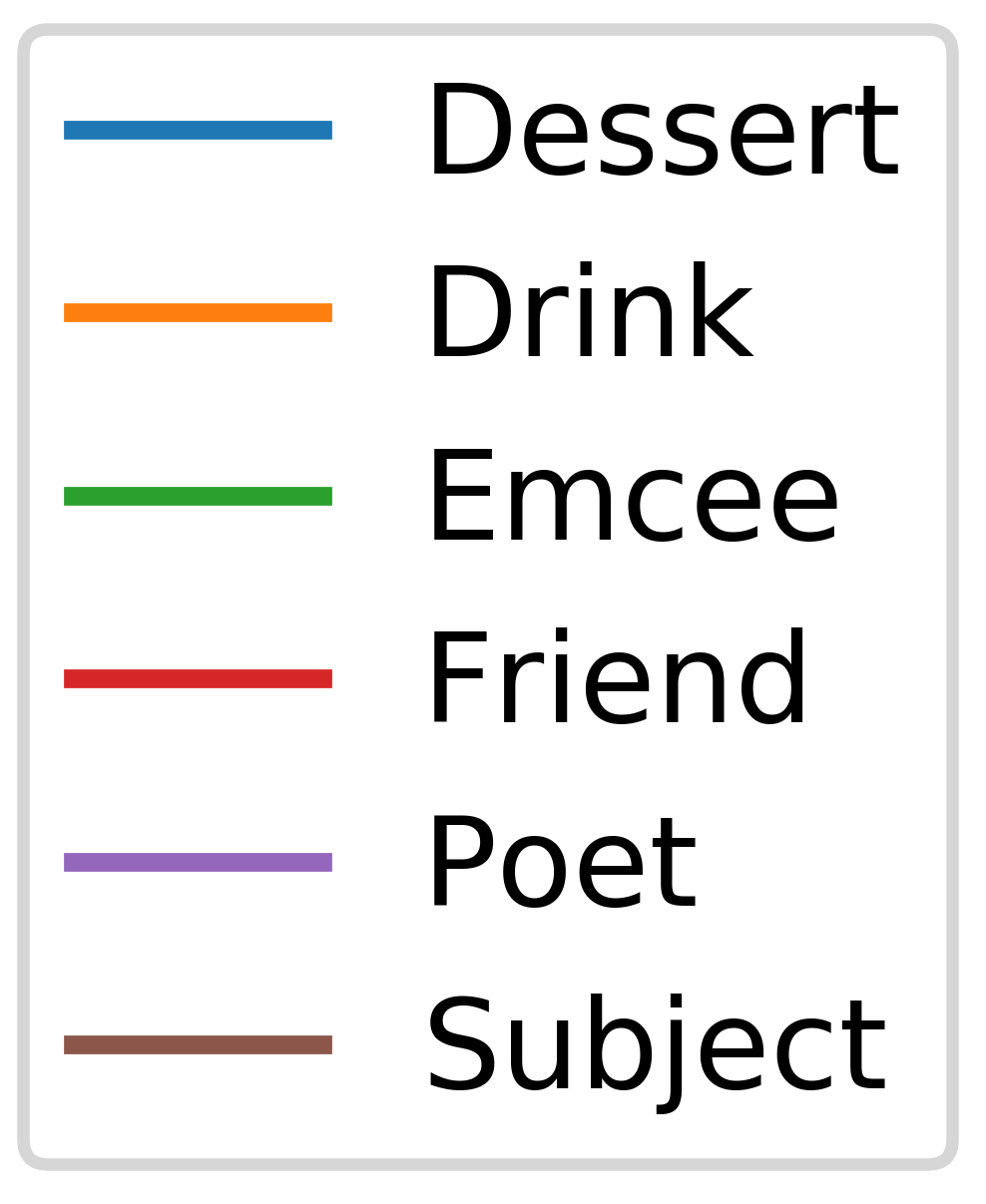}}
\end{subfigure}
\caption{\textbf{Overall and Query-Split Accuracies for Experiment 3.} The chance rate is $1.3\%$, bars denote mean accuracies, and error bars denote maximum and minimum accuracies over three trials. Full learning curves are available in the Supplemental Material.}
\label{fig:Unlimitedtrain_barchart}
\end{figure}

As we show in Figure \ref{fig:Unlimitedtrain_barchart_overall}, all architectures reach above-chance test accuracy, showing that all architectures are sufficient for some amount of generalization to previously unseen fillers, when given a training set with an unlimited pool of fillers.

The different degrees of success shown in Figure \ref{fig:Unlimitedtrain_barchart_overall} reflect the uneven difficulty of queries. In Figure \ref{fig:Unlimitedtrain_barchart_split}, we show network performance separated by query. The RNN does not succeed in answering any role query, and the LSTM succeeds only in answering queries of the \textit{Subject} role. The Fast Weights and reduced NTM networks learn to perform role-filler binding for all role queries. As indicated by the schema structure in Figure \ref{fig:story_paths}, the filler corresponding to the \textit{Subject} role is easiest to identify, as it always occurs at the same location in the story. In contrast, all other roles have variable locations within the story, depending on the probabilistically chosen sequence of story states. 

The results of this experiment show that if networks are provided an unlimited pool of training fillers, then some learn to perform role-filler binding, while simpler networks either fail to learn role-filler binding, or learn to perform role-filler binding for only the simplest role queries.

\subsection*{Decoding Analysis}
\begin{figure}[!htb]
\centering
\begin{subfigure}[b]{0.49\linewidth}
{\includegraphics[width = \linewidth]{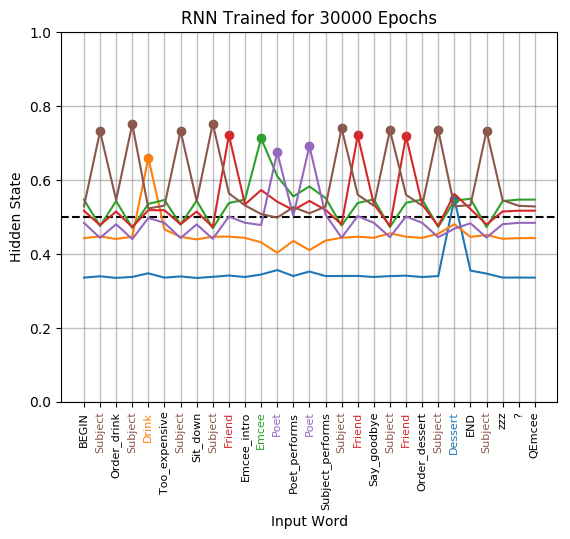}}
\caption{RNN.}
\label{fig:decoding_rnn}
\end{subfigure}
\begin{subfigure}[b]{0.49\linewidth}
{\includegraphics[width = \linewidth]{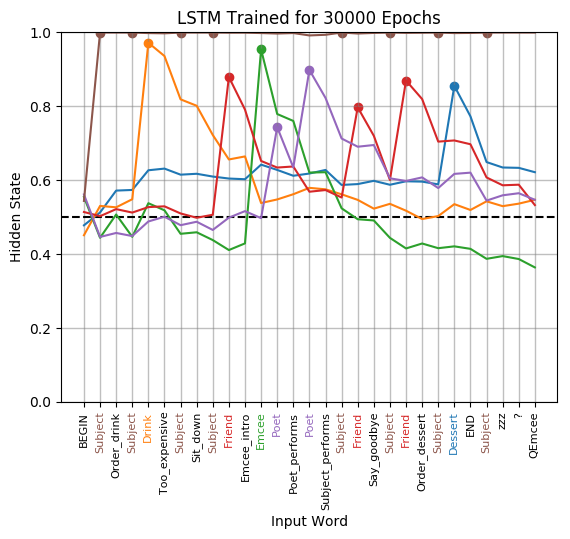}}
\caption{LSTM.}
\label{fig:decoding_lstm}
\end{subfigure}
\begin{subfigure}[b]{0.49\linewidth}
{\includegraphics[width = \linewidth]{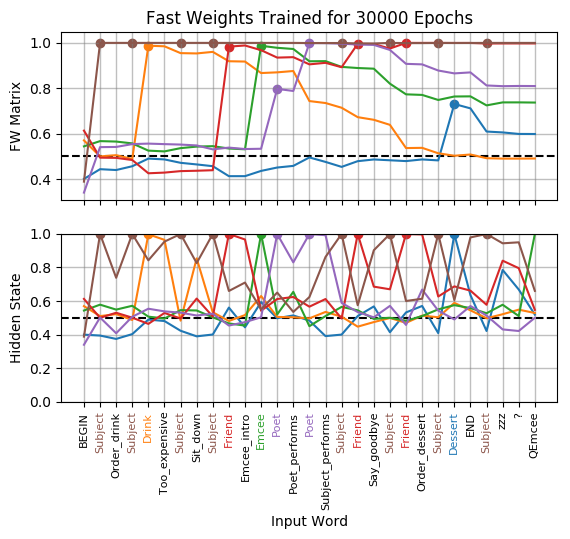}}
\caption{Fast Weights.}
\label{fig:decoding_fw}
\end{subfigure}
\begin{subfigure}[b]{0.49\linewidth}
{\includegraphics[width = \linewidth]{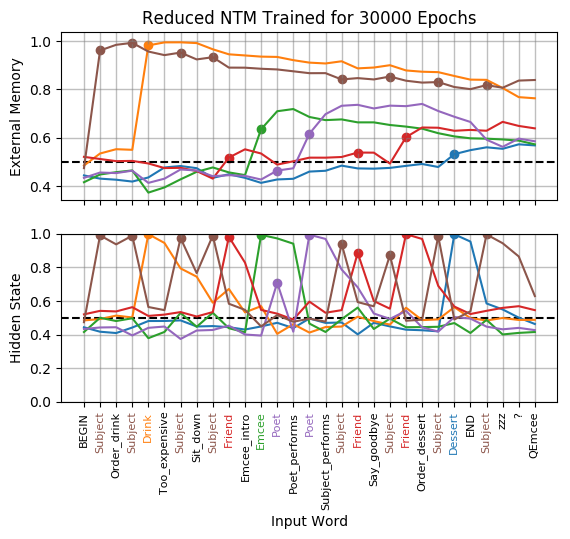}}
\caption{Reduced NTM.}
\label{fig:decoding_ntm2}
\end{subfigure}
\begin{subfigure}[b]{\linewidth}
{\includegraphics[width = \linewidth]{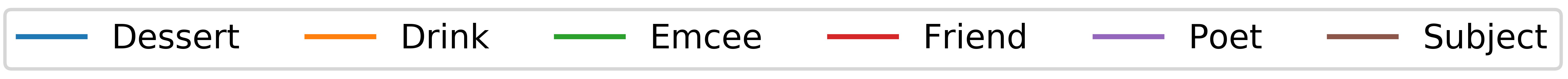}}
\end{subfigure}\vspace{5mm}
\caption{\textbf{Decoding Scores for Experiment 3.}
For the RNN (subfigure a), which is unable to solve any of the tasks, decoding scores for each of the task words are around the chance rate at the end of the input sequence. The LSTM (subfigure b), which solves only the \textit{QSubject} task, only maintains the ability to decode the \textit{Subject} throughout the input sequence.
The Fast Weights and reduced NTM architectures show a similar trend in the hidden internal state of the controllers: decoding scores of the hidden states peak when the networks receive the respective filler in the input sequence, then decline as the network receives more words (subfigures c and d, bottom). In comparison, the decoding scores using the external memory (i.e. the fast weights matrix and the NTM's external memory buffer) increase when the network receives the corresponding filler in its input and the scores remain high throughout the input sequence (subfigures c and d, top).
The chance rate is $50\%$.}
\label{fig:decoding_allQs}
\end{figure}

We performed decoding analyses on the four networks trained in Experiment $3$, to gain insight into how and if the memory components aid in learning role-filler binding. We find that the ability to decode correct fillers, at the time that the network is presented with the query, corresponds to networks' success in role-filler binding. We show the ranking scores in Figure \ref{fig:decoding_allQs}.

The RNN is unable to solve any of the six tasks (as shown in Figure \ref{fig:Unlimitedtrain_barchart_split}), and its decoding scores hover around the chance rate ($50\%$) for all tasks, as we show in Figure \ref{fig:decoding_rnn}.
From the LSTM's hidden state we could decode only the \textit{Subject} at an above-chance rate at query-time, mirroring this network's ability to only solve \textit{QSubject} tasks (Figure \ref{fig:decoding_lstm}).

With the Fast Weights architecture, we decoded using either the controller's hidden state or the set of associative fast weights. We show these decoding scores in Figure \ref{fig:decoding_fw}. The decoding scores from the controller's hidden state mirror those of the LSTM network's hidden state: The scores peak when the network receives the filler in its input, then decline as the network receives more words. (An exception is decoding scores for the \textit{Subject} filler. This could be due to the fixed location of the \textit{Subject} filler, and the non-fixed locations of the other fillers.) We see this trend regardless of whether the network was trained to retrieve a certain filler or not. In contrast, the decoding scores using the Fast Weights matrix increase when the network receives the corresponding filler in its input and remain above chance at query-time.

We see a similar pattern between standard and external memory components with the reduced NTM (Figure \ref{fig:decoding_ntm2}), where the decoding scores using the reduced NTM's controller's hidden state mirror those of the LSTM, and the decoding scores from the reduced NTM's external memory matrix mirror those of the Fast Weights matrix. These results suggest that networks learn to solve tasks by storing the relevant information using their external memory components (either the external memory buffer or the fast weights matrix), while the controller acts as a conduit to receive these words and move them to the external memory component.

\begin{figure}[!htb]
\centering
\begin{subfigure}[b]{0.49\linewidth}
{\includegraphics[width = \linewidth]{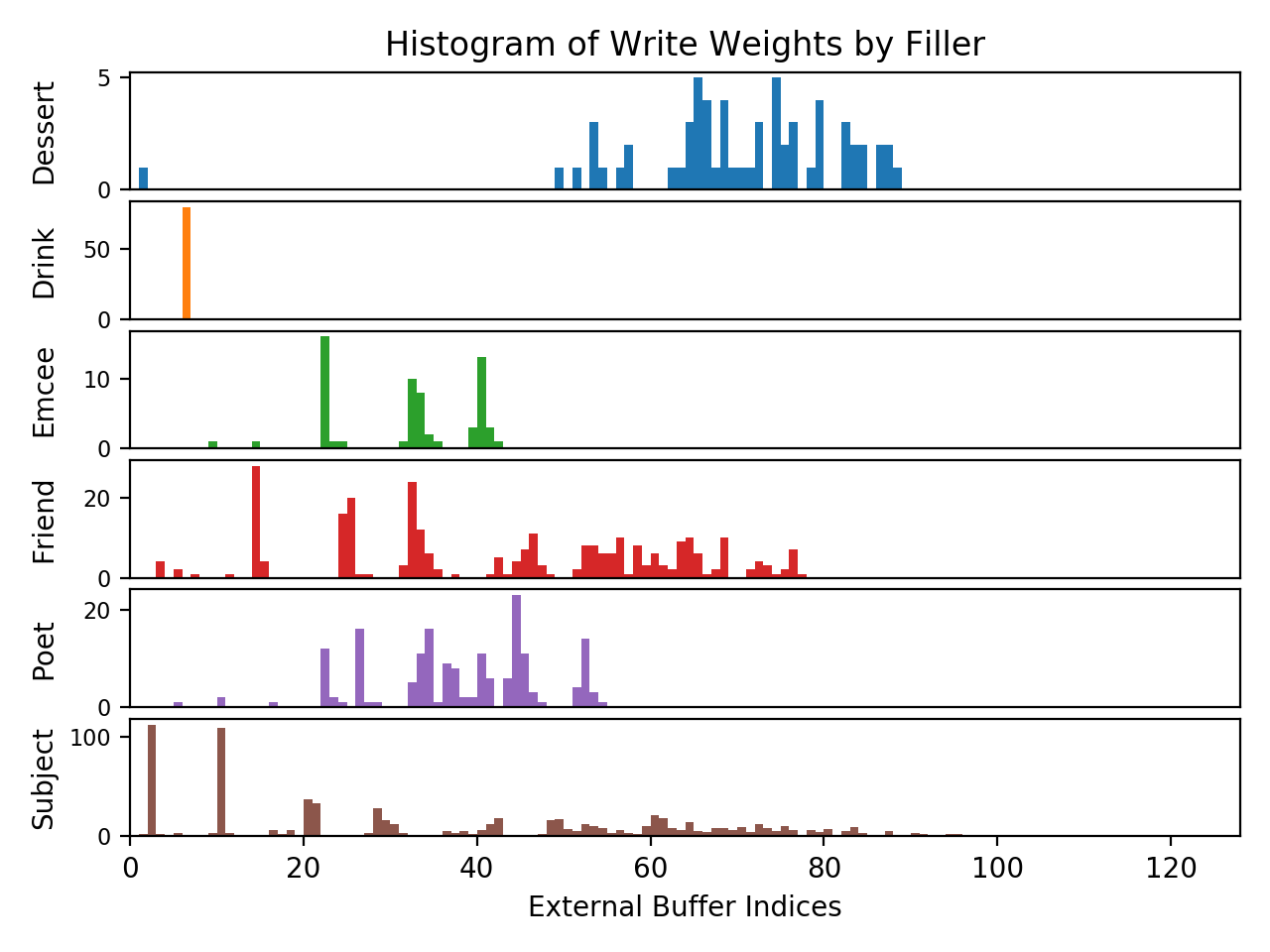}}
\caption{Maximum Write Weights.}
\end{subfigure}
\begin{subfigure}[b]{0.49\linewidth}
{\includegraphics[width = \linewidth]{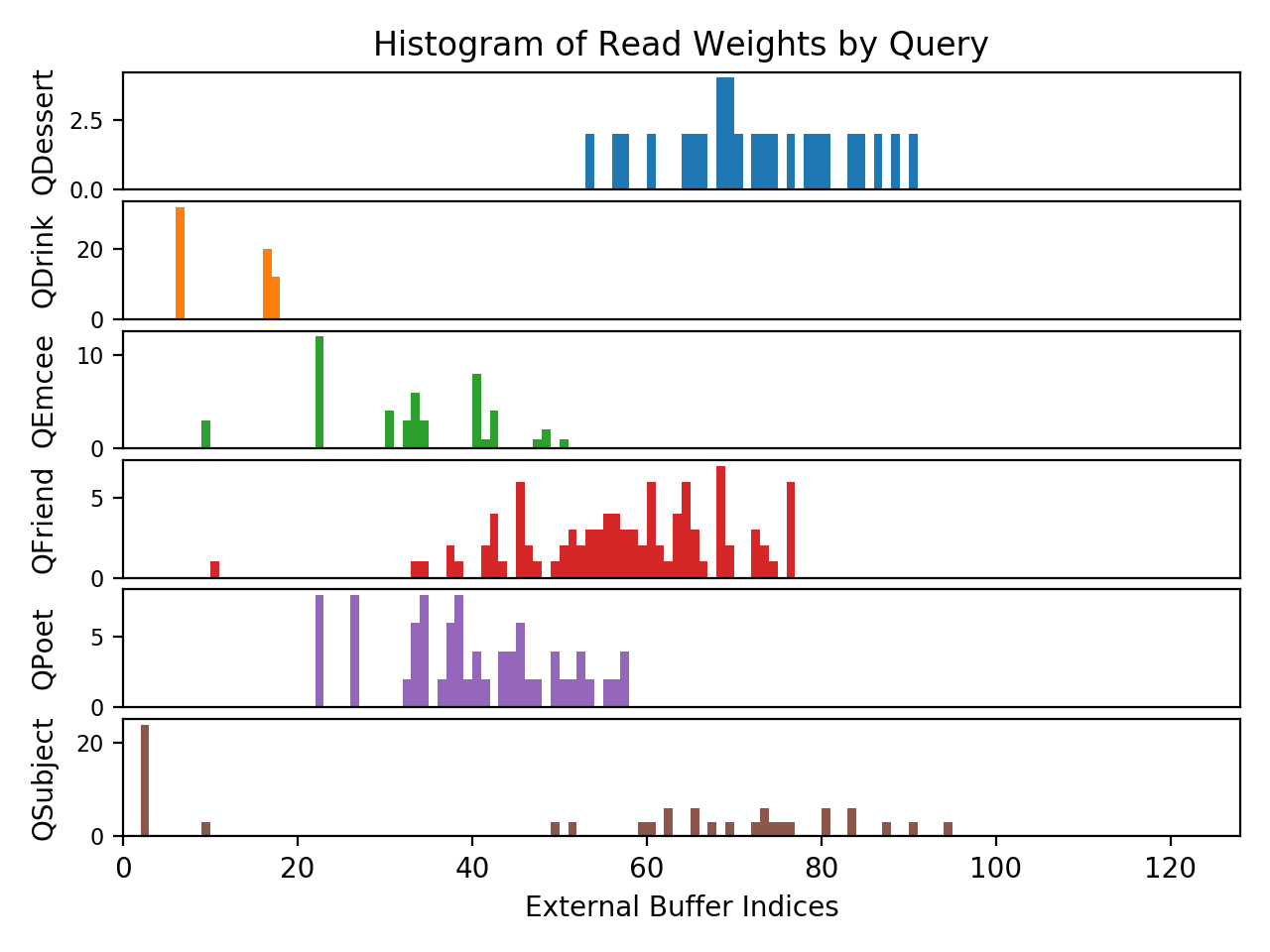}}
\caption{Maximum Read Weights.}
\end{subfigure}
\begin{subfigure}[b]{0.49\linewidth}
{\includegraphics[width = \linewidth]{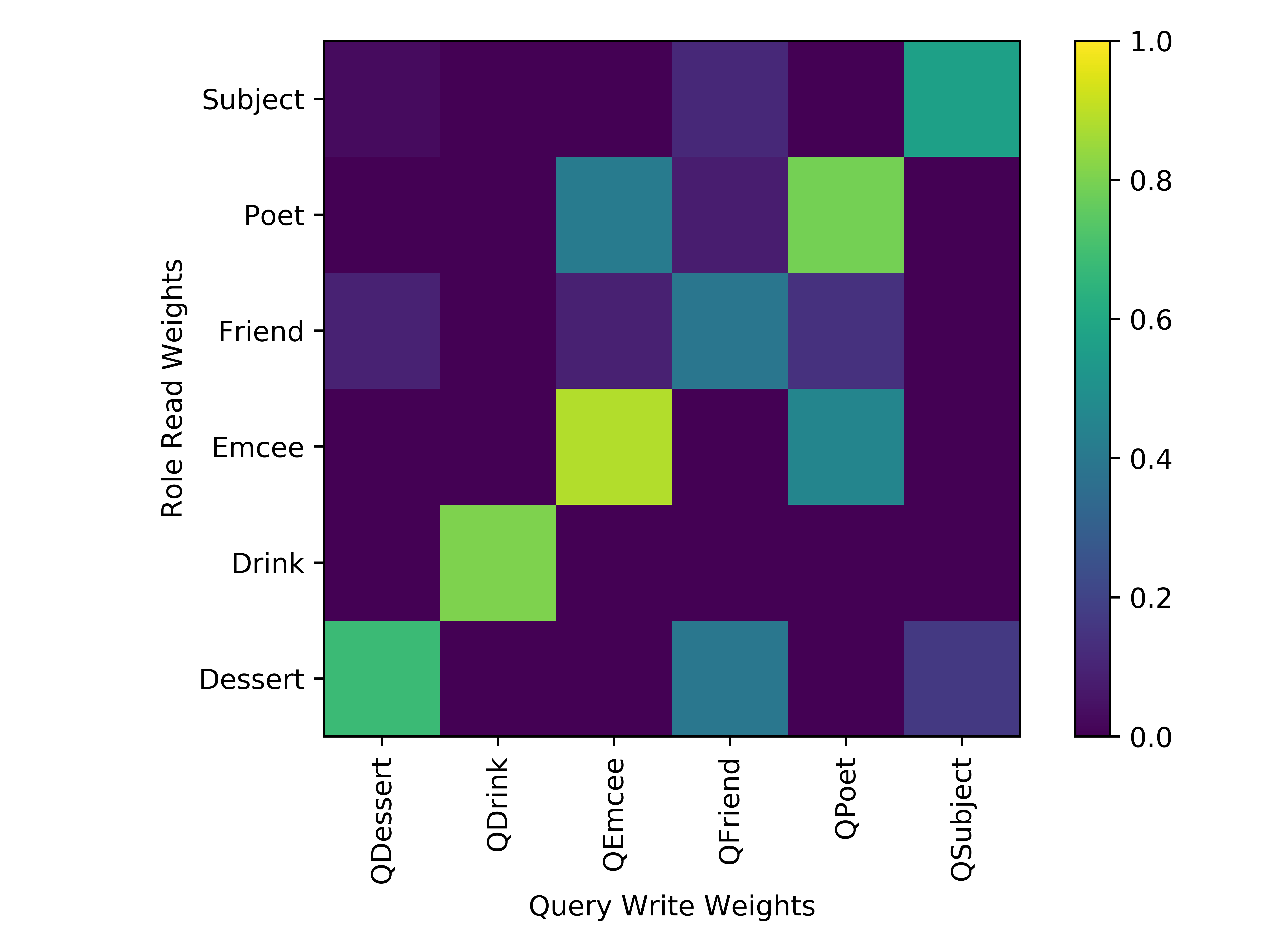}}
\centering
\caption{Pearson Correlations Between Distributions of Maximum Read and Write Weights.}
\end{subfigure}
\caption{\textbf{Maximum Read and Write Weights for Reduced NTM External Memory Buffer.} Read and write weights indicate that networks use a location-based strategy. For all examples in the test set of Experiment 3, we determined the index of the external memory buffer with the highest read or write weights, at chosen input timesteps. The network associates different external memory slots more strongly with each role. Correspondence between write weights when the network receives the filler for role ``X", and read weights when the network is queried for role ``X" indicate that networks use a location-based strategy to save and retrieve role-filler pairs. (a) shows histograms of maximum write weights, for input timesteps in which the network receives the filler corresponding to the role denoted by y-axis labels. (b) shows histograms of maximum read weights (for all four read heads), for each input timestep in which the network receives the query denoted by y-axis labels. (c) shows Pearson correlations between the distribution of maximum write indices when the network receives fillers corresponding to each role (on the x-axis), and the distribution of maximum read indices when the network receives queries asking for each role (on the y-axis).}
\label{fig:ntm2weights}
\end{figure}

Furthermore, the read and write weights of the reduced NTM indicate that the network learns to store and retrieve role-filler bindings using a location-based strategy. The read weights influence where in the external memory buffer the network should read from, and the write weights influence where in the external memory buffer the network should write to. In Figure \ref{fig:ntm2weights} we show the maximum write weights at input timesteps corresponding to the appearance of fillers corresponding to each role, and the maximum read weights at the timestep at which the network makes its prediction. The rows showing network write weights show that the network associates different slots in memory with each role. Furthermore, these results show that the distribution of network read weights when asked to predict query ``X" correspond to the network write weights when the filler corresponding to ``X" occurs in the input sequence, suggesting that the network saves and retrieves role-filler pairs using a location-based strategy.

These findings suggest that networks that learned to perform role-filler binding learned to store relevant information in external memory components.

\subsection*{Correlation Violation Tests}
\begin{figure}[htb!]
\centering
\includegraphics[width=0.7\linewidth]{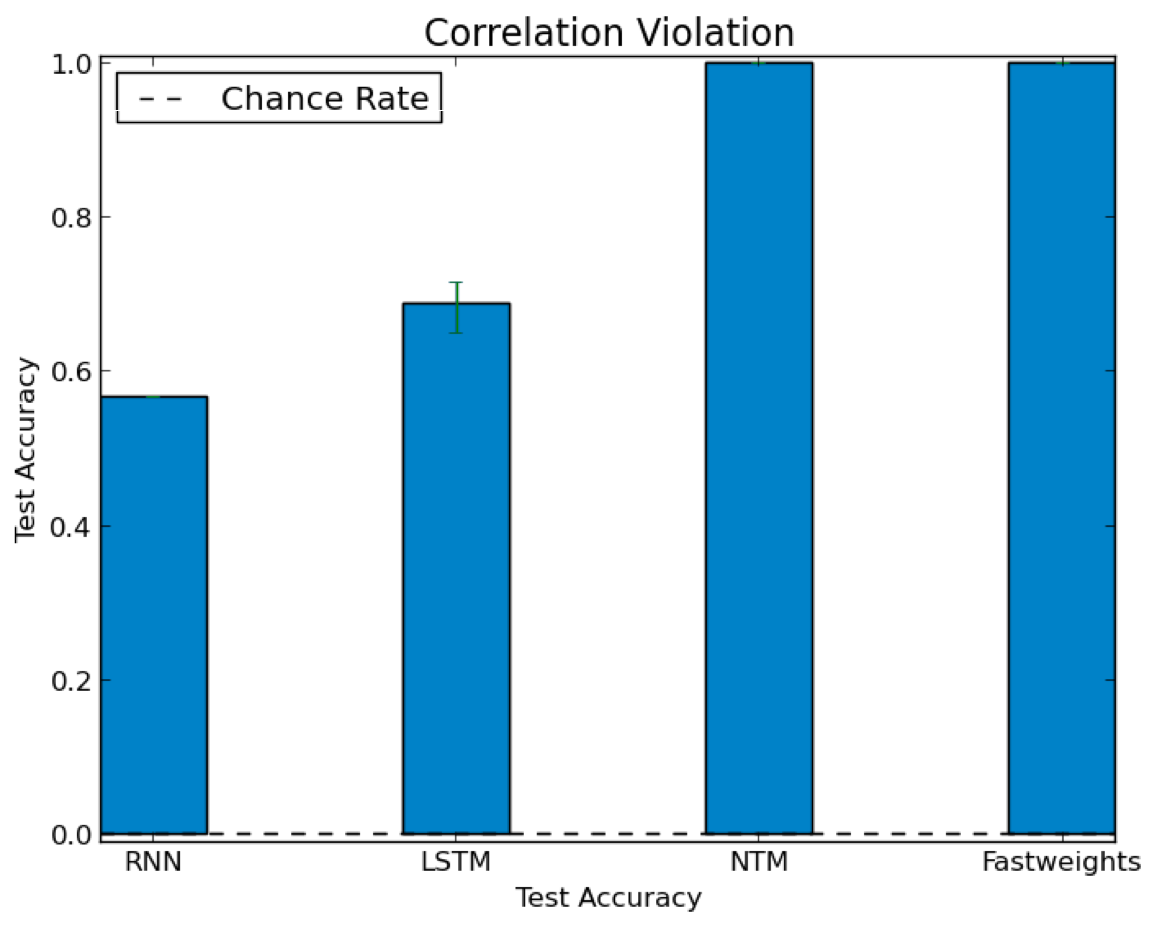}
\caption{\textbf{Test Accuracy for Correlation Violation.} Some networks learn to perform role-filler binding, even when test examples break role-filler correlation observed during training. The chance accuracy rate is 0.02\%, bars denote mean accuracies, and error bars denote maximum and minimum accuracies over three trials.}\label{fig:correlation_violation}
\end{figure}

In our test of correlation violation, we trained networks using stories with strong role-filler correlations. We constructed these correlations by excluding each filler from a specific role during training. We then constructed a test set that deliberately broke these correlations seen during training. In the test set, we associated each role only with fillers that were excluded from that role during training. When networks receive inputs with role-filler bindings that violate correlations seen during training, they accurately extract the correlation-violating role-filler bindings. All networks reach above chance accuracy on this test set (the NTM and Fast Weights architectures reach $100\%$ accuracy), as we show in Figure \ref{fig:correlation_violation}, showing that the networks can generalize to role-filler bindings that violate statistics seen during training.

We further showed that a network can retain information about training statistics, even though they are able to generalize to fillers that violate correlations seen during training. We focus on the reduced NTM in this section, because it was shown to perform well in previous tests and our analysis of its read and write weights suggested that it could both retain information about training statistics and generalize to previously unseen fillers.

We trained networks in two different settings. In the first setting, all word vectors seen during training were generated by independently drawing each index of the word vector from a $N(0,1)$ distribution and then normalizing the vector to have unit Euclidean norm. In the second setting, all word vectors seen during training were initially generated as in setting one. Then, for $50\%$ of the train fillers we added $0.5$ to the even indices. Then, we gave these networks inputs consisting solely of nonsense words, followed by one of the query words, and recorded the vectors predicted by the networks.

To test the network's retention of filler statistics seen during training, we tested whether the network in the second setting defaults to predicting fillers with higher values in the even indices, compared to the predicted odd indices. We trained $25$ networks in each setting, and recorded each network's prediction in response to each of the $4$ query words, resulting in $100$ predictions per network type. We then tested whether networks in these settings learned to predict greater even indices. For each predicted vector, we computed the two-sample t-statistic comparing the even and odd indices of the predicted vector, resulting in $100$ t-statistics per training setting. Then for each setting, we used these $100$ t-statistics in a 1-sample t-test comparing the mean of these t-statistics to $0$.

For networks trained in the first setting, in which both the even and odd indices were drawn from a $N(0,1)$ distribution, the 1-sample t-test produced a p-value of $0.12$, meaning that these vectors did not predict significantly different even and odd indices. For networks trained in the second setting, in which $50\%$ of fillers seen during training had even indices drawn from a $N(0.5, 1)$ distribution, the 1-sample t-test produced a p-value of $0.000013$, showing that these networks predicted even indices that were significantly different from odd indices. Note that adding the statistical regularity in the second setting did not adversely affect the NTM's ability to recall randomly-generated fillers -- performance was still at 100\%.
Putting all of these findings together, networks in the second setting were able to generalize to examples that violated correlations seen during training, and retain correlation statistics seen during training, which they defaulted to when given ambiguous input examples.

No networks consistently succeed in performing role-filler binding with shuffled story test sets.

\section*{Discussion}
In previous work, models successfully performed role-filler binding when given explicitly labeled inputs \citep{Kriete:2013,Elman:2019,Franklin:2019} or the same fillers during train and test \citep{StJohnMcClelland:1990,Miikkulainen:1991,Hinaut:2013}. They failed when given test stories that included previously unseen fillers that violated correlations seen during training \citep{Puebla:2019}. In our experiments, we find that some models can perform role-filler binding on inputs that are not explicitly labeled with role-filler pairs. Moreover, they do so on novel fillers, and on bindings that violate correlations seen during training. At the same time, these models show clear sensitivity to statistical regularities in the fillers when asked to respond to ambiguous prompts at test.

In our experiments we found that networks trained on stories including a small pool of train fillers (as in Puebla et al. 2019) failed at generalizing to novel fillers, while the networks that both had external memory and were trained on a larger pool of fillers succeeded. This suggests that networks must see a sufficiently diverse set of fillers during training, in order to learn to separate the representations of roles and fillers. Furthermore, the successful networks, which were not tested in the Puebla et al. (2019) study, contained external memory (the set of fast weights or the NTM's external memory buffer). Our findings suggest that external memory is an important architectural component for networks to learn role-filler binding. Previous work has emphasized the importance of preserving role-filler independence by separating the representations of bindings and fillers, and has suggested that conjunctive coding (which would encode the role-filler binding by storing a modified representation of the filler) fails to preserve this independence \citep{FodorPylyshyn:1988,Hummel:2004}. We hypothesize that the Fast Weights and reduced NTM models' external memory components allow them to preserve this independence.

Our decoding analyses suggest that the successful networks learn to store fillers in their external memory. The ability to decode specific fillers from certain networks, and the correspondence between decoding success and network performance on role-filler binding tasks, give insight into when and where the networks store role-filler bindings. For instance, we were able to decode fillers at query-time from Fast Weights' and NTM's external memory components (but not from the controller); this suggests that these networks perform role-filler binding by storing bindings in the external memory components, and then retrieving the correct binding upon receiving the query. The read and write weight distributions of the reduced NTM suggest that the network learns to use certain locations in memory to store fillers corresponding to each role. External memory provides a separate storage space, which gives the model a way to encode a filler's role by storing it in a certain location in external memory, without needing to encode role-filler bindings in a modified representation of the filler. We note that our results show models sufficient for performing this task, but we do \textit{not} conclude that all RNN and LSTM networks are unable to solve this task. For instance, future work with larger LSTM models might show that models with larger network capacities could somehow also learn role-filler binding, without explicitly provided external memory.

Our experiments include bindings that violate correlations seen during training, and that include fillers that were not seen during training. This means that, in order to successfully perform our tasks at test time, networks must have learned to store role-filler pairs in a way that was not specific to role-filler correlations seeing during training. Furthermore, our experiments used randomly generated vectors (rather than pre-trained word vectors) to represent input words, meaning that networks could not rely on pre-encoded linguistic knowledge.

Networks fail when given stories presented in a different order, showing that they need some amount of shared structural similarity between train and test examples, in order to identify and store role-filler pairs. Future work could explore training regimes that use shuffled stories during training, to see whether this could make models robust to test stories with shuffled segments. Future work could also test how the loss of certain architectural components affects the ability to perform schema-learning, by lesioning certain components of an artificial network. Our experiments indicated that the diversity of training examples influences whether networks learn to perform role-filler binding. Future work could investigate how much a pool of train fillers needs to be expanded to allow networks to learn a representation that extends to previously unseen fillers. Furthermore, the psychology literature indicates that schemata can encode biases and stereotypes, affecting how humans interpret new information and recall previous information \citep{Bartlett:1932}. Future work could explore whether and how models adapt to changes in the underlying schema over the course of training, and how examples provided during training might translate into biases encoded in schemata, and how these principles apply in situations where multiple schemata need to be learned \citep{Franklin:2019}. The impact of external memory components in our experiments suggests that episodic memory (a cognitive ability that corresponds to the external memory components) may play an important role in applying schemata to novel fillers, and further work could investigate this connection.

\section*{Conclusions}
Our experiments find networks that can perform role-filler binding, where we define role-filler binding as the ability to bind arbitrary fillers to certain roles without receiving explicitly labeled bindings, even when these pairings violate correlations seen during training. Successful models could generalize to previously unseen fillers when given external memory and when given novel fillers during training. These models could perform bindings with novel fillers, and with fillers that violate role-filler correlations seen during training. This suggests that the networks learn to extract the relations from the structure of the input, rather than relying on role-filler correlations seen during training.

Previous work has suggested that indirection can support variable bindings \citep{Kriete:2013}, and Complementary Learning Systems theory hypothesizes that the coordination of two distinct systems allows models to both learn overall structures shared between experiences and rapidly learn new items \citep{McClelland:1995}. While future work could explore whether networks without external memory could also perform these tasks, our experiments show that, at the very least, there are models with external memory that learn role-filler binding. These findings provide a possible mechanism for connectionist architectures to learn role-filler binding relations, which are a central component for learning flexible, structured cognitive representations.
\bibliography{root}

\section*{Supplemental Material}

\subsection*{Learning Curves.}\label{sec:learning_curves}
In this section we include learning curves corresponding to accuracies depicted in previously presented bar plots. In each plot we show the mean accuracy, with error ribbons for maximum and minimum accuracies, over three trials.

\begin{figure}[!htb]
\centering
\begin{subfigure}{0.5\linewidth}
{\includegraphics[width = \linewidth]{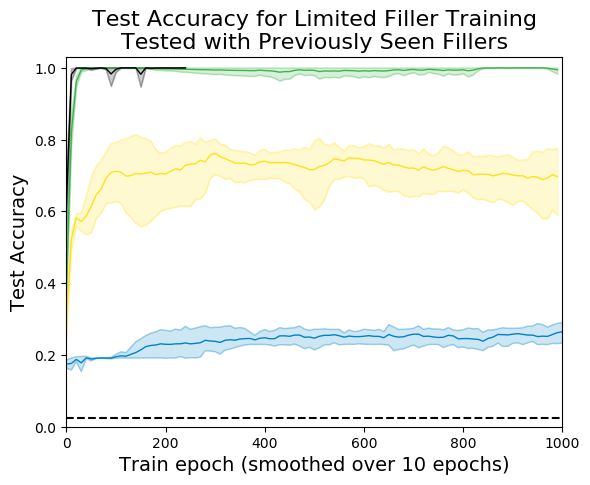}}
\end{subfigure}
\begin{subfigure}{0.1\linewidth}
{\includegraphics[width = \linewidth]{\figurepath/decoding_legend.png}}
\end{subfigure}
\caption{Test accuracy for Experiment $1$. (Limited Filler Training Tested with Previously Seen Fillers, originally shown in Figure 2. The chance accuracy rate is $2.3\%$.}
\end{figure}

\begin{figure}[!htb]
\centering
\begin{subfigure}{0.4\linewidth}
{\includegraphics[width = \linewidth]{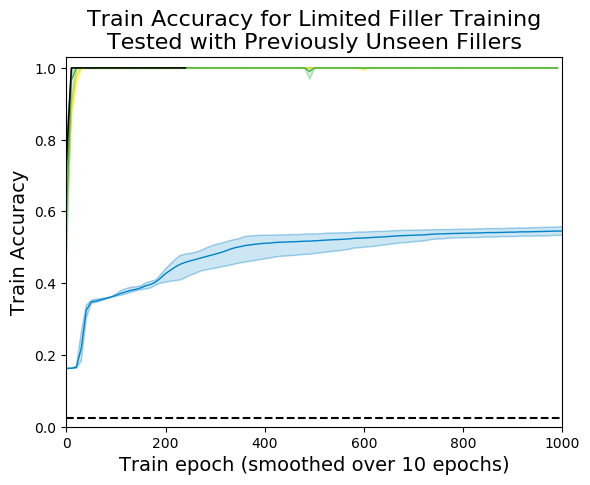}}
\caption{}
\end{subfigure}
\begin{subfigure}{0.4\linewidth}
{\includegraphics[width = \linewidth]
{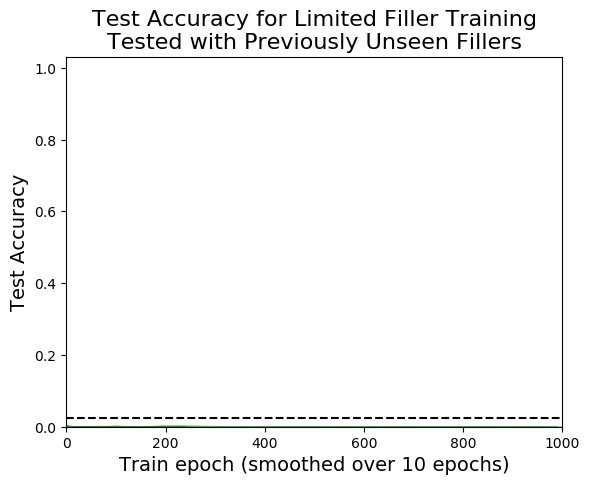}}
\caption{}
\end{subfigure}
\begin{subfigure}{0.1\linewidth}
{\includegraphics[width = \linewidth]{\figurepath/decoding_legend.png}}
\end{subfigure}
\caption{Train and test accuracy for Experiment $2$. (Limited Filler Training Tested with Previously Unseen Fillers, originally shown in Figure 3. The chance accuracy rate is $2.3\%$.}
\end{figure}

\begin{figure}[!htb]
\centering
\begin{subfigure}{0.4\linewidth}
{\includegraphics[width = \linewidth]{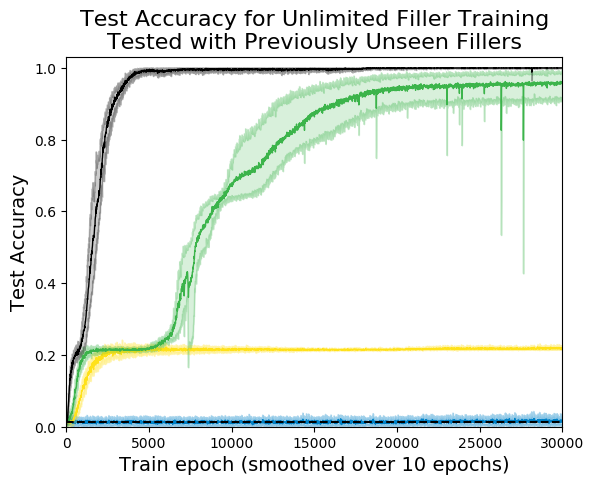}}
\caption{}
\end{subfigure}
\begin{subfigure}{0.4\linewidth}
{\includegraphics[width = \linewidth]
{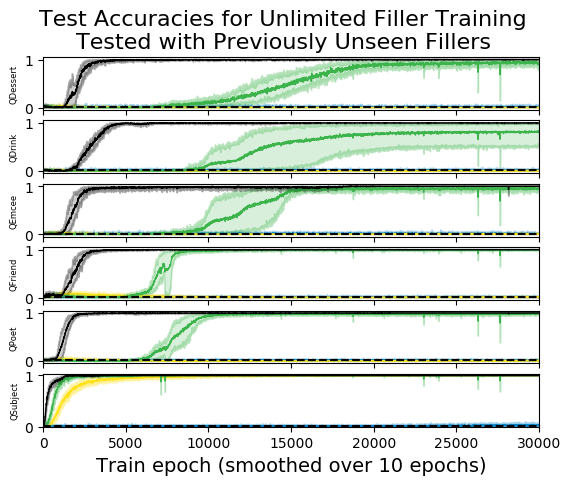}}
\caption{}
\end{subfigure}
\begin{subfigure}{0.1\linewidth}
{\includegraphics[width = \linewidth]{\figurepath/decoding_legend.png}}
\end{subfigure}
\caption{Overall (left) and Query-Split (right) accuracy for Experiment $3$. (Unlimited Filler Training Tested with Previously Unseen Fillers, originally shown in Figure 4. The chance rate is $1.3\%$.}
\end{figure}
\end{document}